\documentclass[sigconf]{acmart}

\AtBeginDocument{%
  \providecommand\BibTeX{{%
    \normalfont B\kern-0.5em{\scshape i\kern-0.25em b}\kern-0.8em\TeX}}}

\usepackage{epsfig}
\usepackage{graphicx}
\usepackage{amsmath}

\usepackage{balance}

\usepackage{booktabs}
\usepackage{multirow}
\usepackage{pifont}
\newcommand{\cmark}{\ding{51}}%
\newcommand{\xmark}{\ding{55}}%

\setcopyright{rightsretained}

\begin{document}
\fancyhead{}

%% Rights management information.  This information is sent to you
%% when you complete the rights form.  These commands have SAMPLE
%% values in them; it is your responsibility as an author to replace
%% the commands and values with those provided to you when you
%% complete the rights form.
\copyrightyear{2021}
\acmYear{2021}
\acmConference[ICDAR '21]{Proceedings of the 2021 Workshop on Intelligent Cross-Data Analysis and Retrieval}{August 21--24, 2021}{Taipei, Taiwan}
\acmBooktitle{Proceedings of the 2021 Workshop on Intelligent Cross-Data Analysis and Retrieval (ICDAR '21), August 21--24, 2021, Taipei, Taiwan}
\acmPrice{}
\acmDOI{10.1145/3463944.3469097}
\acmISBN{978-1-4503-8529-9/21/08}

\title{ST-HOI: A Spatial-Temporal Baseline for \\ Human-Object Interaction Detection in Videos}

\author{Meng-Jiun Chiou}
\authornote{The work was done during a research internship at ASUS Intelligent Cloud Services.}
\affiliation{%
    \institution{National University of Singapore}
    \city{Singapore}
    \country{Singapore}
}
\email{mengjiun.chiou@u.nus.edu}

\author{Chun-Yu	Liao}
\affiliation{
    \institution{ASUS Intelligent Cloud Services}
    \city{Taipei}
    \country{Taiwan}
}
\email{mist_liao@asus.com}
% \authornote{Corresponding author}

\author{Li-Wei Wang}
\affiliation{
    \institution{ASUS Intelligent Cloud Services}
    \city{Taipei}
    \country{Taiwan}
}
\email{popo55668@gmail.com}

\author{Roger Zimmermann}
\affiliation{
    \institution{National University of Singapore}
    \city{Singapore}
    \country{Singapore}
}
\email{rogerz@comp.nus.edu.sg}

\author{Jiashi Feng}
\affiliation{
    \institution{National University of Singapore}
    \city{Singapore}
    \country{Singapore}
}
\email{elefjia@nus.edu.sg}

% \renewcommand{\shortauthors}{Anonymous Submission \#262}

\begin{abstract}
   Detecting human-object interactions (HOI) is an important step toward a comprehensive visual understanding of machines. 
   While detecting non-temporal HOIs (\emph{e.g.}, \textit{sitting on} a chair) from static images is feasible, it is unlikely even for humans to guess temporal-related HOIs (\emph{e.g.}, \textit{opening}/\textit{closing} a door) from a single video frame, where the neighboring frames play an essential role.
   However, conventional HOI methods operating on only static images have been used to predict temporal-related interactions, which is essentially guessing without temporal contexts and may lead to sub-optimal performance.
   In this paper, we bridge this gap by detecting video-based HOIs with explicit temporal information.
   We first show that a naive temporal-aware variant of a common action detection baseline does not work on video-based HOIs due to a feature-inconsistency issue. 
   We then propose a simple yet effective architecture named Spatial-Temporal HOI Detection (ST-HOI) utilizing temporal information such as human and object trajectories, correctly-localized visual features, and spatial-temporal masking pose features. 
   We construct a new video HOI benchmark dubbed VidHOI\footnote{The dataset and source code are available at \url{https://github.com/coldmanck/VidHOI}} where our proposed approach serves as a solid baseline.
\end{abstract}





\maketitle

\section{Introduction}
\label{sec:introduction}

\begin{figure}[t!]
\centering
\includegraphics[width=.95\columnwidth]{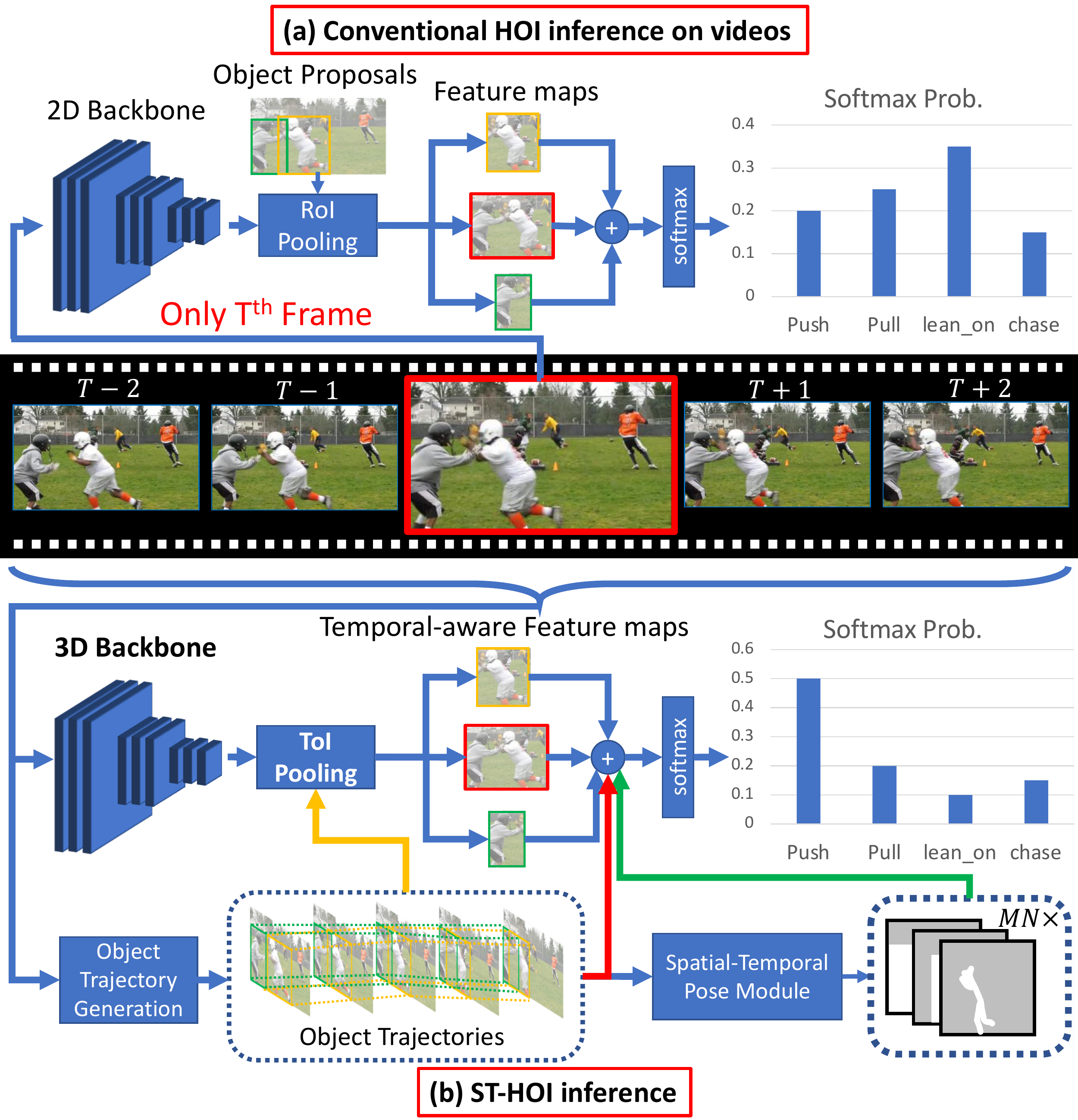}
\vspace{-1.2em}
\caption{An illustrative comparison between conventional HOI methods and our ST-HOI when inferencing on videos. 
(a) Traditional HOI approaches (\emph{e.g.}, the baseline in \cite{wan2019pose}) take in only the target frame and predict HOIs based on ROI-pooled visual features.
% , detect object proposals with an off-the-shelf object detector and perform region-of-interest pooling (RoIPool) on CNN features to obtain human/object and context features, which are then be used to predict an interaction. 
These models are unable to differentiate between {\fontfamily{qcr}\selectfont push},  {\fontfamily{qcr}\selectfont pull} or {\fontfamily{qcr}\selectfont lean on} in this example due to the lack of temporal context.
(b) ST-HOI takes in not only the target frame but neighboring frames and exploits temporal context based on trajectories. 
% A 3D-CNN \cite{feichtenhofer2019slowfast} is adopted to extract the temporal feature across the video segment, on which tube-of-interest pooling (ToIPool) \cite{hou2017tube} is applied to extract temporal-aware visual features. 
ST-HOI can thus differentiate temporal-related interactions and prefers {\fontfamily{qcr}\selectfont push} to other interactions in this example.
}
\vspace{-1.3em}
\label{fig:motivation}
\end{figure}

\begin{figure*}[t!]
\centering
\includegraphics[width=0.93\textwidth]{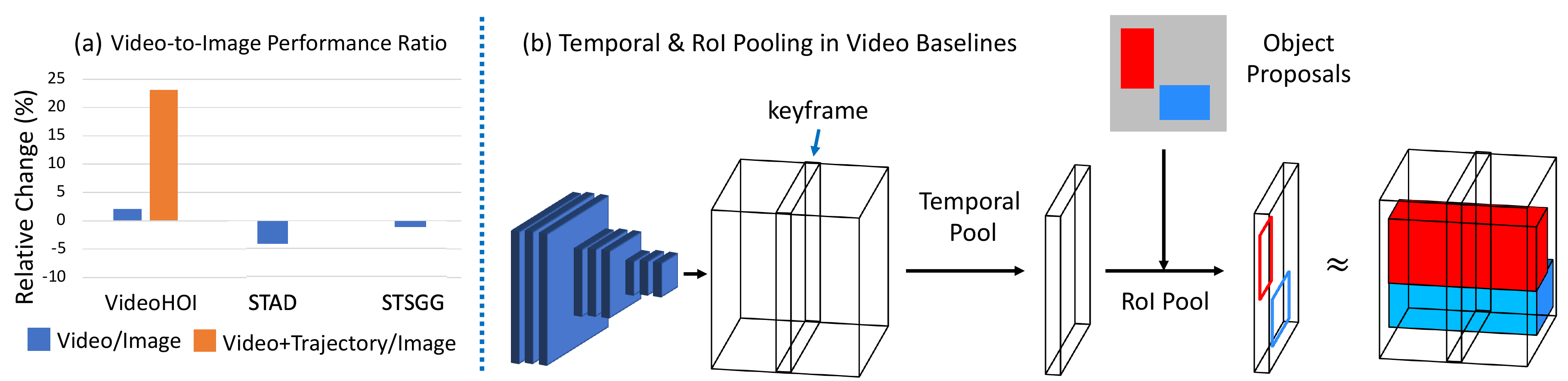}
\vspace{-0.7em}
\caption{(a) Relative performance change (in percentage), on different video tasks by replacing 2D-CNN backbones with 3D ones (blue bars) \cite{tran2015learning,carreira2017quo,feichtenhofer2019slowfast}, and on VideoHOI by adding trajectory feature (tangerine bar).
\textbf{VideoHOI} (in triplet mAP) is to detect HOI in videos and was performed ourselves on our VidHOI benchmark.
\textbf{STAD} \cite{gu2018ava} (in triplet mAP) means Spatial-Temporal Action Detection and was performed on AVA dataset \cite{gu2018ava}. 
\textbf{STSGG} \cite{ji2020action} (PredCls mode; in Recall@20) stands for Spatial-Temporal Scene Graph Generation and was performed on Action Genome \cite{ji2020action}. 
(b) An illustration of temporal-RoI pooling in 3D baselines (\emph{e.g.} \cite{feichtenhofer2019slowfast}). 
Temporal pooling is usually applied to the output of the penultimate layer of a 3D-CNN (shape of $d \times T \times H \times W$) which average-pools along the time axis into shape of $d \times 1 \times H \times W$, followed by RoI Pooling to obtain feature maps of shape $d \times 1 \times h \times w$.
This temporal-RoI pooling, however, is equivalent to pooling the instance-of-interest feature at the same location in the keyframe throughout the video segment, which is erroneous for moving humans and objects.
}
\vspace{-0.5em}
\label{fig:image_video_baseline_comparison}
\end{figure*}

Thanks to the rapid development of deep learning \cite{Goodfellow-et-al-2016,he2016deep}, machines are already surpassing or approaching human level performance in language tasks \cite{wu2016google}, acoustic tasks \cite{yin2019multi}, and vision tasks (\emph{e.g.,} image classification \cite{7410480} and visual place recognition \cite{chiou2020zero}).
Researchers thus started to focus on how to replicate these successes to other semantically higher-level vision tasks (\emph{e.g.}, visual relationship detection \cite{lu2016visual,chiou2021visual}) and vision-language tasks (\emph{e.g.}, image captioning \cite{vinyals2015show} and visual question answering \cite{antol2015vqa}) so that machines learn not just to recognize the objects but to \textit{understand} their relationships and the contexts.
Especially, human-object interaction (HOI) \cite{smith2013gaze,chao2018learning,gao2018ican,qi2018learning,li2019transferable,wang2019deep,wan2019pose,gupta2019no,li2020pastanet,wang2020learning,liao2020ppdm,ulutan2020vsgnet,zhong2020polysemy} aiming to detect actions and spatial relations among humans and salient objects in images/videos has attracted increasing attention, as we sometimes task machines to understand human behaviors, \emph{e.g.,} pedestrian detection \cite{zhang2017towards} and unmanned store systems \cite{su11184965}.

Although there are abundant studies that have achieved success in detecting HOI in static images, the fact that few of them \cite{jain2016structural,qi2018learning,sunkesula2020lighten} consider temporal information (\emph{i.e.}, neighboring frames before/after the target frame) when performed on video data means they are actually ``guessing'' temporal-related HOIs with only naive co-occurrence statistics.
While conventional image-based HOI methods (\emph{e.g.}, the baseline model in \cite{wan2019pose}) can be used for inference on videos, they treat input frames as independent and identically distributed (\textit{i.i.d.}) data and make independent predictions for neighboring frames. 
However, video data are sequential and structured by nature and thus are not \textit{i.i.d.}
What is worse is that, without temporal context these methods are unable to differentiate (especially, opposite) temporal interactions, such as {\fontfamily{qcr}\selectfont push} versus {\fontfamily{qcr}\selectfont pull} a human and {\fontfamily{qcr}\selectfont open} versus {\fontfamily{qcr}\selectfont close} a door.
% While these methods still work for certain interactions like spatial relationships (\emph{e.g.} {\fontfamily{qcr}\selectfont above}, {\fontfamily{qcr}\selectfont next to}), 
As shown in Figure \ref{fig:motivation}(a), given a video segment, traditional HOI models operate on a single frame at a time and make predictions based on 2D-CNN (\emph{e.g.}, \cite{he2016deep}) visual features.
% perform region-of-interest (RoI) pooling on feature maps extracted from a CNN \cite{he2016deep} with object proposals to obtain human/object features, which are then fused and classified by linear layers.
These models by nature could not distinguish interactions between two people such as {\fontfamily{qcr}\selectfont push}, {\fontfamily{qcr}\selectfont pull}, {\fontfamily{qcr}\selectfont lean on} and {\fontfamily{qcr}\selectfont chase}, which are visually similar in static images. 
A possible reason causing video-based HOI underexplored is the lack of a suitable video-based benchmark and a feasible setting.
To bridge this gap, we first construct a video HOI benchmark from VidOR \cite{shang2019annotating}, dubbed \textbf{VidHOI}
% \footnote{VidHOI will be released to facilitate researches in video-based HOI.}
, where we follow the common protocol in video and HOI tasks to use a keyframe-centered strategy.
% \footnote{More details on VidHOI are given in section \ref{subsec:dataset}}
With VidHOI, we urge the use of video data and propose \textbf{VideoHOI} as -- in both training and inference -- performing HOI detection with videos. % By using video data, as illustrated in Figure \ref{fig:motivation}(b) we take in $T$ frames of each video segment (e.g. $T=32$ frames sampled from each $\sim$2s segment for 30-fps videos) centered at the target frame (\emph{e.g.} the $16^{\text{th}}$ frame). 

Spatial-Temporal Action Detection (STAD) is another task bearing a resemblance to VideoHOI by requiring to localize the human and detect the actions being performed in videos. 
Note that STAD does not consider the objects that a human is interacting with.
STAD is usually tackled by first using a 3D-CNN \cite{tran2015learning,carreira2017quo} as the backbone to encode temporal information into feature maps.
This is followed by RoI pooling with object proposals to obtain actor features, which are then classified by linear layers. 
Essentially, this approach is similar to a common HOI baseline illustrated in Figure \ref{fig:motivation}(a) and differs only in the use of 3D backbones and the absence of interacting objects.
Based on conventional HOI and STAD methods, a naive yet intuitive idea arises: \textit{can we enjoy the best of both worlds, by replacing 2D backbones with 3D ones and exploiting visual features of interacting objects?}
This idea, however, did not work straightforwardly in our preliminary experiment, where we replaced the backbone in the 2D baseline \cite{wan2019pose} with the 3D one (\emph{e.g.}, SlowFast \cite{feichtenhofer2019slowfast}) to perform VideoHOI.
% Due to the lack of video-based benchmark for HOI detection, 
% For all the three experiments, the only difference between image- and video-models is the backbone (2D- or 3D-CNN, where 3D-CNN is an inflated version of 2D-CNN). 
The relative change of performance after replacing the backbone is presented in the left most entry in Figure \ref{fig:image_video_baseline_comparison}(a) with a blue bar. 
In VideoHOI experiment, the 3D baseline provides only a limited relative improvement ($\sim$2\%), which is far from satisfactory considering the additional temporal context.
In fact, this phenomenon has also been observed in two existing works under similar settings \cite{gu2018ava,ji2020action}, where both experiments in STAD and another video task Spatial-Temporal Scene Graph Generation (STSGG) present an even worse, counter-intuitive result: replacing the backbone is actually harmful (also presented as blue bars in Figure \ref{fig:image_video_baseline_comparison}(a)).
We probed the underlying reason by analyzing the architecture of these 3D baselines and found that, surprisingly, temporal pooling together with RoI pooling does not work reasonably.
As illustrated in Figure \ref{fig:image_video_baseline_comparison}(b), temporal pooling followed by RoI pooling, which is a common practice in conventional STAD methods, is equivalent to cropping features of the same region across the whole video segment without considering the way objects move.
It is not unusual for moving humans and objects in neighboring frames to be absent from its location in the target keyframe. 
Temporal-and-RoI-pooling features at the same location could be getting erroneous features such as other humans/objects or meaningless background.
Dealing with this inconsistency, we propose to recover the missing spatial-temporal information in VideoHOI by considering human and object trajectories. 
% Refer to the red arrow in Figure \ref{fig:motivation}(b) for an illustration of this additional branch.
The performance change of this temporal-augmented 3D baseline on VideoHOI is represented by the tangerine bar in Figure \ref{fig:image_video_baseline_comparison}(a), where it achieves $\sim$23\% improvement, in sharp contrast to $\sim$2\% of the original 3D baseline.
This experiment reveals the importance of incorporating the "correctly-localized" temporal information.

Keeping the aforementioned ideas in mind, in this paper we propose \textbf{S}patial-\textbf{T}emporal baseline for \textbf{H}uman-\textbf{O}bject \textbf{I}nteraction detection in videos, or \textbf{ST-HOI}, which makes accurate HOI prediction with instance-wise spatial-temporal features based on trajectories. As illustrated in Figure \ref{fig:motivation}(b), three kinds of such features are exploited in ST-HOI: (a) trajectory features (moving bounding boxes; shown as the red arrow), (b) correctly-localized visual features (shown as the yellow arrow), and (c) spatial-temporal actor poses (shown as the green arrow). 

The contribution of our work is three-fold. First, we are among the first to identify the feature inconsistency issue existing in the naive 3D models which we address with simple yet “correct” spatial-temporal feature pooling.
% , as demonstrated in the preliminary experiment.
Second, we propose a spatial-temporal model which utilizes correctly-localized visual features, per-frame box coordinates and a novel, temporal-aware masking pose module to effectively detect video-based HOIs.
% VideoHOI with promising results. 
Third, we establish the keyframe-based VidHOI benchmark to motivate research in detecting spatial-temporal aware interactions and hopefully inspire VideoHOI approaches utilizing the multi-modality data, \emph{i.e.}, video frames, texts (semantic object/relation labels) and audios.

% Due to the unavailability of a feasible dataset and a proper evaluation setting for VideoHOI, we adopt keyframe-centered evaluation strategy and establish a VideoHOI benchmark named VidHOI, where we show ST-HOI is superior to a conventional HOI method (2D baseline) as well as a naive 3D baseline.

%%% Related Work %%%
\section{Related Work}
\label{sec:related_work}

\subsection{Human-Object Interaction (HOI)}
\label{subsec:hoi}
HOI Detection aims to reason over interactions between humans (actors) and target objects. 
HOI is closely related to visual relationship detection \cite{lu2016visual,chiou2021visual} and scene graph generation \cite{xu2017scene}, in which the subject in \textit{(subject-predicate-object)} are not restricted to a human. 
HOI in static images has been intensively studied recently \cite{smith2013gaze,chao2018learning,gao2018ican,qi2018learning,li2019transferable,wang2019deep,wan2019pose,gupta2019no,li2020pastanet,wang2020learning,liao2020ppdm,ulutan2020vsgnet,zhong2020polysemy}. 
Most of the existing methods can be divided into two categories by the order of human-object pair proposal and interaction classification. 
The first group \cite{gao2018ican,chao2018learning,li2019transferable,wang2019deep,wan2019pose,gupta2019no,ulutan2020vsgnet,li2020pastanet} performs human-object pair generation followed by interaction classification, while the second group \cite{gkioxari2018detecting,qi2018learning,wang2020learning,liao2020ppdm} first predicts the most probable interactions performed by a person followed by associating them with the most-likely objects.
Our ST-HOI belongs to the first group as we establish a temporal model based on trajectories (continuous object proposals).

In contrast to the popularity of image-based HOI, there are only a few of studies in VideoHOI \cite{koppula2015anticipating,jain2016structural,qi2018learning,sunkesula2020lighten} and, to the best of our knowledge, all of which conducted experiments on CAD-120 \cite{koppula2013learning} dataset.
In CAD-120, the interactions are defined by merely 10 high-level activities (\emph{e.g.}, {\fontfamily{qcr}\selectfont making cereal} or {\fontfamily{qcr}\selectfont microwaving food}) in 120 RGB-D videos. % , of which the length ranges from 22 to 150 frames, \emph{i.e.} less than five seconds for the longest videos.
This setting is not favorable to real-life scenarios where machines may be asked to understand more fine-grained actions.
Moreover, previous methods \cite{koppula2015anticipating,jain2016structural,qi2018learning} adopted pre-computed hand-crafted features such as SIFT \cite{lowe2004distinctive} which have been outperformed by deep neural networks, and ground truth features including 3D poses and depth information from RGB-D videos which are unlikely to be available in real life scenarios.
While \cite{sunkesula2020lighten} adopted a ResNet \cite{he2016deep} as their backbone, their method is inefficient by requiring $M \times N$ computation for extracting $M$ humans' and $N$ objects' features. 
% In addition, their model still utilizes ground truth bounding boxes. % and is trained in a two-stage manner, leading to sub-optimal performance. 
Different from these existing methods, we evaluate on a larger and more diversified video HOI benchmark dubbed VidHOI, which includes annotations of 50 predicates on thousands of videos. 
We then propose a spatial-temporal HOI baseline that operates on RGB videos and does not utilize any additional information.

\subsection{Spatial-Temporal Action Detection (STAD)}
\label{subsec:stad}
% \subsubsection{Spatial-Temporal Action Detection (STAD)}
STAD aims to localize actors and detect the associated actions (without considering interacting objects). 
One of the most popular benchmark for STAD is AVA \cite{gu2018ava}, where the annotation is done at a sampling frequency of 1 Hz and the performance is measured by framewise mean AP. 
We followed this annotation and evaluation style when constructing VidHOI, where we converted the original labels into the same format. 

As explained in section \ref{sec:introduction}, a standard approach to STAD \cite{tran2015learning,carreira2017quo} is extracting spatial-temporal feature maps with a 3D-CNN followed by RoI pooling to crop human features, which are then classified by linear layers. 
As shown in Figure \ref{fig:image_video_baseline_comparison}(a), a naive modification that incorporates RoI-pooled human/object features does not work for VideoHOI.
In contrast, our ST-HOI tackles VideoHOI by incorporating multiple temporal features including trajectories, correctly-localized visual features and spatial-temporal masking pose features.

% Recent studies in STAD focus on the ways to encode spatial-temporal features, \emph{e.g.} with a better backbone \cite{feichtenhofer2019slowfast}, memory feature bank \cite{wu2019long} or self-attention \cite{wang2018non}.

\begin{figure*}[t!]
\centering
\includegraphics[width=0.93\textwidth]{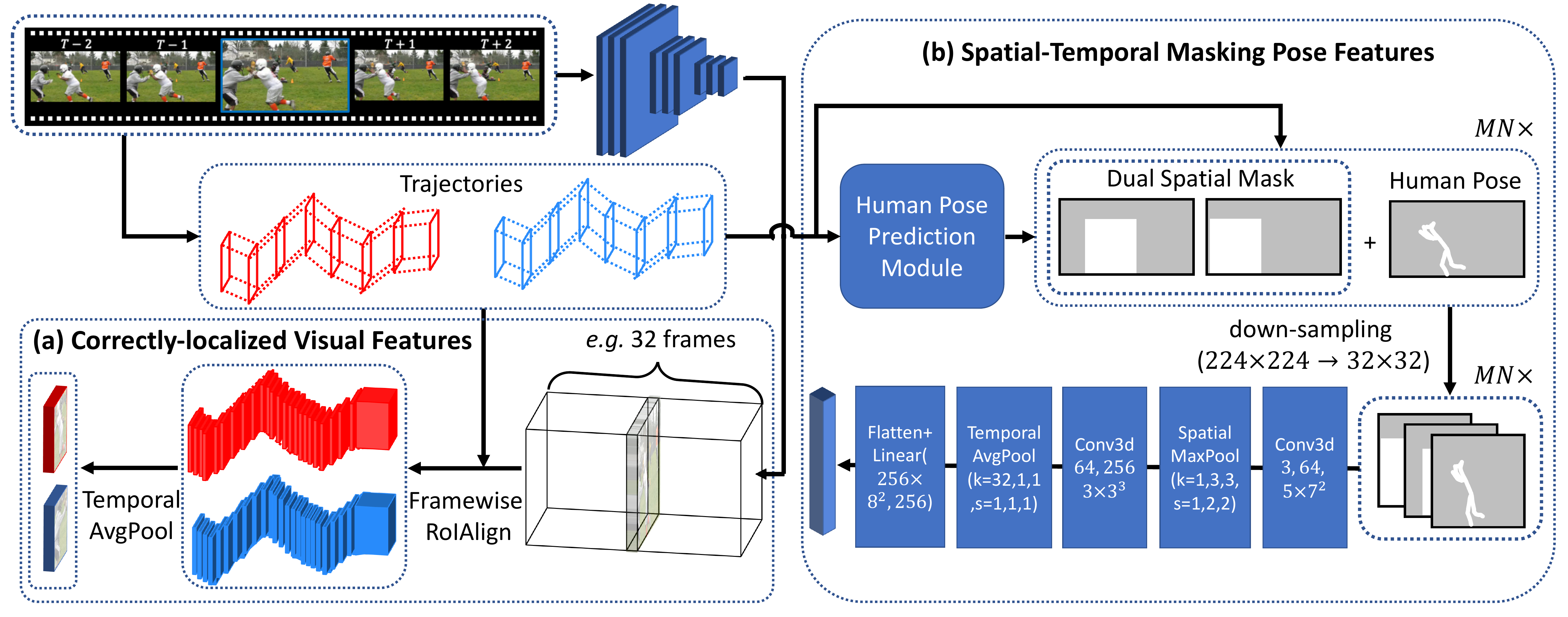}
\vspace{-1em}
\caption{An illustration of the two proposed spatial-temporal features.
% The procedure of generating correctly-lolcalized visual features with ToIPool \cite{hou2017tube}. 
(a) In contrast to performing RoI pooling followed by temporal pooling like \cite{wu2019long,feichtenhofer2019slowfast}, we adopt a reverse approach to first frame-wise RoI-pool instance feature maps using trajectories, which are then averaged pool along the time axis to get correctly-localized visual features. 
(b) With $N$ object trajectories (including $M$ human), for each frame we utilize a trained human pose prediction model (\emph{e.g.}, \cite{fang2017rmpe}) to generate 2D actor pose feature and extract a dual spatial mask for all $M \times (N-1)$ valid pair. 
The pose feature and the mask are concatenated and down-sampled, followed by two 3D convolution layers and spatial-temporal pooling to generate the masking pose features.
}
\label{fig:toipool_and_pose}
\vspace{-0.5em}
\end{figure*}

\subsection{Spatial-Temporal Scene Graph Generation}
\label{subsec:stsgg}

Spatial-Temporal Scene Graph Generation (STSGG) \cite{ji2020action} aims to generate symbolic graphs representing pairwise visual relationships in video frames.
A new benchmark, Action Genome, is also proposed in \cite{ji2020action} to facilitate researches in STSGG.
Ji et al. \cite{ji2020action} dealt with STSGG by combining off-the-shelf scene graph generation models with long-term feature bank \cite{wu2019long} on top of a 2D- or 3D-CNN, where they found that the 3D-CNN actually undermines the performance.
While observing similar results in VidHOI (Figure \ref{fig:image_video_baseline_comparison}(a)), we go one step further to find out the underlying reason is that RoI features across frames were erroneously pooled.
We correct this by utilizing object trajectories and applying Tube-of-Interest (ToI) pooling on generated trajectories to obtain correctly-localized position information and feature maps throughout video segments.

% Different from vision tasks on static-images, video understanding puts emphasis on reasoning contents across a sequence of frames, where video models usually take in video segments (\emph{e.g.} 5-60 frames). 
% Great efforts have been made on video understanding tasks including action classification, temporal action localization and spatial-temporal action detection (STAD).

%%% Methodology %%%
\section{Methodology}
\label{sec:methodology}

\subsection{Overview}
\label{subsec:overview}
We follow STAD approaches \cite{tran2015learning,carreira2017quo,feichtenhofer2019slowfast} to detect VideoHOI in a keyframe-centric strategy. 
Denote $V$ as a video which has $T$ keyframes with sampling frequency of 1 Hz as $\{I_t\}, t=\{1,...,T\}$, and denote $C$ as the number of pre-defined interaction classes.
Given $N$ instance trajectories including $M$ human trajectories ($M \leq N$) in a video segment centered at the target frame, for human $m \in \{1,...,M\}$ and object $n \in \{1,...,N\}$ in keyframe $I_t$, we aim to detect pairwise human-object interactions $r_t = \{0,1\}^C$, where each entry $r_{t,c}, c \in \{1,...,C\}$ means whether the interaction $c$ exists or not.

Refer to Figure \ref{fig:motivation}(b) for an illustration of our ST-HOI.
Our model takes in a video segment (sequence of $T$ frames) centered at $I_t$ and utilizes a 3D-CNN as the backbone to extract spatial-temporal feature maps of the whole segment.
To rectify the mismatch caused by temporal-RoI pooling, based on $N$ object (including human) trajectories $\{j_i\}, i=\{1,..,N\}, j_i \in \mathbb{R}^{T \times 4}$ we generate temporal-aware features including correctly-localized features and spatial-temporal masking pose features.
These features together with trajectories are concatenated and classified by linear layers.
Note that we aim at a simple but effective temporal-aware baseline to VideoHOI so that we do not utilize tricks in STAD such as non-local block \cite{wang2018non} or long-term feature bank \cite{wu2019long}, and in image-based HOI like interactiveness \cite{li2019transferable}, though we note that these may be used to boost the performance.
% The model makes prediction on all valid pairs (where subject is {\fontfamily{qcr}\selectfont human}).

\subsection{Correctly-localized Visual Features}
\label{subsec:temporal_visual_features}

We have discussed in previous sections on inappropriately pooled RoI features.
We propose to tackle this issue by reversing the order of temporal pooling and RoI-pooling.
This approach has recently been proposed in \cite{hou2017tube} and named as tube-of-interest pooling (ToIPool).
Refer to Figure \ref{fig:toipool_and_pose}(a) for an illustration.
Denote $v \in \mathbb{R}^{d \times T \times H \times W}$ as the output of the penultimate layer of our 3D-CNN backbone, and denote $v_t \in \mathbb{R}^{d \times H \times W}$ as the $t$-th feature map along the time axis.
Recall that we have $N$ trajectories centered at a keyframe.
Following the conventional way, we also exploit visual context when predicting an interaction, which is done by utilizing the union bounding box feature of a human and an object.
For example, the sky between {\fontfamily{qcr}\selectfont human} and {\fontfamily{qcr}\selectfont kite} could help to infer the correct interaction {\fontfamily{qcr}\selectfont fly}.
% For $M$ human trajecories ($M<N$), there are $M \times N$ possible pairs, and thus we append $M \times N$ union-box trajectories to $\{j_i\},  i=\{1,...,N,N+1,...,N+M\times N\}$.
Recall that $j_i$ represents the trajectory of object $i$, where we further denote $j_{i,t}$ as the 2D bounding box at time $t$.
The spatial-temporal instance features $\{\bar{v}_i\}$ are then obtained using ToIPool with RoIAlign \cite{he2017mask} by % \{\bar{v}_i\}, i=\{1,..,N, N+1, ..., N+M\times N\}
\begin{equation}
     \bar{v}_i = \frac{1}{T} \sum_{t=1}^{T} \text{RoIAlign}(v_{t}, j_{i,t}),
\end{equation}
where $\bar{v}_i \in \mathbb{R}^{d \times h \times w}$ and $h$ and $w$ means height and width of the pooled feature maps, respectively. 
$\bar{v}_i$ is flattened before concatenating with other features.

\subsection{Spatial-Temporal Masking Pose Features}
\label{subsec:spaital_temporal_human_poses}

Human poses have been widely utilized in image-based HOI methods \cite{li2019transferable,gupta2019no,wan2019pose} to exploit characteristic actor pose to infer some special actions.
In addition, some existing works \cite{wang2019deep,wan2019pose} found that spatial information can be used to identify interactions.
For instance, for {\fontfamily{qcr}\selectfont human-ride-horse} one can imagine the actor's skeleton as legs widely open (on horse sides), and the bounding box center of {\fontfamily{qcr}\selectfont human} is usually on top of that of {\fontfamily{qcr}\selectfont horse}.
However, none of the existing works consider this mechanism in temporal domain: when riding a horse the {\fontfamily{qcr}\selectfont human} should be moving with {\fontfamily{qcr}\selectfont horse} as a whole.
We argue that this temporality is an important property and should be utilized as well.
% To the best of our knowledge, we are the first to extend the mechanism into temporal domain by designing with 3D-CNNs and temporal pooling.
% We are the first work to exploit this spatial-temporal masking feature.

The spatial-temporal masking pose module is presented at Figure \ref{fig:toipool_and_pose}(b).
Given $M$ human trajectories, we first generate $M$ spatial-temporal pose features with a trained human pose prediction model.
On frame $t$, the predicted human pose $h_{i,t} \in \mathbb{R}^{17 \times 2}, i=\{1,..,M\}, t=\{1,..,T\}$ is defined as 17 joint points mapped to the original image.
We transform $h_{i,t}$ into a skeleton on a binary mask with $f_h: \{h_{i,t}\} \in \mathbb{R}^{17 \times 2} \to \{\bar{h}_{i,t}\} \in \mathbb{R}^{1 \times H \times W}$, by connecting the joints using lines, where each line has a distinct value $x \in [0, 1]$. 
This helps the model to recognize and differentiate different poses.

For each of $M \times (N-1)$ valid human-object pairs on frame $t$, we also generate two spatial masks $s_{i,t} \in \mathbb{R}^{2 \times H \times W}, i=\{1,...,M \times (N-1)\}$ corresponding to human and object respectively, where the values inside of each bounding box are ones and outsides are zeroed-out.
These masks enable our model to predict HOI with reference to important spatial information.

For each pair, we concatenate the skeleton mask $\bar{h}_{i,t}$ and spatial masks $s_{i,t}$ along the first dimension to get the initial spatial masking pose feature $p_{i,t} \in \mathbb{R}^{3 \times H \times W}$:
\begin{equation}
    p_{i,t} = [s_{i,t}; \bar{h}_{i,t}].
\end{equation}
We then down-sample $\{p_{i, t}\}$, feed into two 3D convolutional layers with spatial and temporal pooling, and flatten to obtain the final spatial-temporal masking pose feature $\{\bar{p}_{i,t}\}$.

\begin{table*}[t!]
\caption{A comparison of our benchmark VidHOI with existing STAD (AVA \cite{gu2018ava}), image-based (HICO-DET \cite{chao2018learning} and V-COCO \cite{gupta2015visual}) and video-based (CAD-120 \cite{koppula2013learning} and Action Genome \cite{ji2020action}) HOI datasets. VidHOI is the only dataset that provides temporal information from video clips and complete multi-person and interacting-object annotations. 
VidHOI also provides the most annotated keyframes and defines the most HOI categories in the existing video datasets. 
$\dagger$Two less categories as we combine {\fontfamily{qcr}\selectfont adult}, {\fontfamily{qcr}\selectfont child} and {\fontfamily{qcr}\selectfont baby} into a single category, {\fontfamily{qcr}\selectfont person}.
}
\vspace{-1em}
\label{tab:dataset_compare}
\resizebox{\linewidth}{!}{%
\begin{tabular}{@{\extracolsep{3pt}}l cc rrrr rrr@{}}
\hline
 \multirow{2}{*}{Dataset} & Video & Localized & Video & \# Videos & \# Annotated & \# Objects & \# Predicate & \# HOI & \# HOI \\ 
%  \cline{5-8}\cline{9-12}
 & dataset? & object? & hours & & images/frames & categories & categories & categories & Instances \\ 
 \hline
HICO-DET \cite{chao2018learning} & \xmark & \cmark & - & - & 47K & 80 & 117 & 600 & 150K \\
V-COCO \cite{gupta2015visual} & \xmark & \cmark & - & - & 10K & 80 & 25 & 259 & 16K \\
\hline
AVA \cite{gu2018ava} & \cmark & \xmark & 108 & 437 & 3.7M & - & 49 & 80 & 1.6M \\
CAD-120 \cite{koppula2013learning} & \cmark & \cmark & 0.57 & 0.5K & 61K & 13 & 6 & 10 & 32K \\
Action Genome \cite{ji2020action} & \cmark & $\bigtriangleup$ & 82 & 10K & 234K & 35 & 25 & 157 & 1.7M \\
\hline
\textbf{VidHOI} & \cmark & \cmark & 70 & 7122 & \textbf{7.3M} & 78$\dagger$ & 50 & \textbf{557} & 755K \\
\hline
\end{tabular}
}
% \vspace{-1em}
\end{table*}

\subsection{Prediction}
We fuse the aforementioned features, including correctly-localized visual features $\bar{v}$, spatial-temporal masking pose features $p$, and instance trajectories $j$ by concatenating them along the last axis
\begin{equation}
    v_{\text{so}} = [\bar{v}_s; \bar{v}_u; \bar{v}_o; j_s; j_o; \bar{p}_{so}], \\
    % r_{\text{so}} = \argmax_{c_i \in C}\ \text{Softmax}(\sigma(v_{\text{so}}))
\end{equation}
where we slightly abuse the notation to denote the subscriptions $s$ as the subject, $o$ as the object and $u$ as their union region.
$v_{\text{so}}$ is then fed into two linear layers with the final output size being the number of interaction classes in the dataset.
Since VideoHOI is essentially a multi-label learning task, we train the model with per-class binary cross entropy loss.

During inference, we follow the heuristics in image-based HOI \cite{chao2018learning} to sort all the possible pairs by their softmax scores and evaluate on only top 100 predictions.

%%% Experiments %%%

\section{Experiments}
\label{sec:experiments}

\subsection{Dataset and Performance Metric}
\label{subsec:dataset}

While we have discussed in section \ref{subsec:hoi} about the problem of lacking a suitable VideoHOI dataset by analyzing CAD-120 \cite{koppula2013learning}, we further explain why Action Genome \cite{ji2020action} is also not a feasible choice here.
First, the authors acknowledged that the dataset is still incomplete and contains incorrect labels \cite{ji_2020}.
Second, Action Genome is produced by annotating Charades \cite{sigurdsson2016hollywood}, which is originally designed for activity classification where each clip contains only one "actor" performing predefined tasks; should any other people show up, there are neither any bounding box nor interaction label about them.
Finally, the videos are purposedly-generated by volunteers, which are rather unnatural. 
In contrast, VidHOI are based on VidOR \cite{shang2019annotating} which is densely annotated with all humans and predefined objects showing up in each frame.
VidOR is also more challenging as the videos are non-volunteering user-generated and thus jittery at times.
A comparison of VidHOI and the existing STAD and HOI datasets is presented in Table \ref{tab:dataset_compare}.

\begin{figure}[t]
\begin{center}
% \fbox{\rule{0pt}{1in} \rule{0.9\linewidth}{0pt}}
\includegraphics[width=\linewidth]{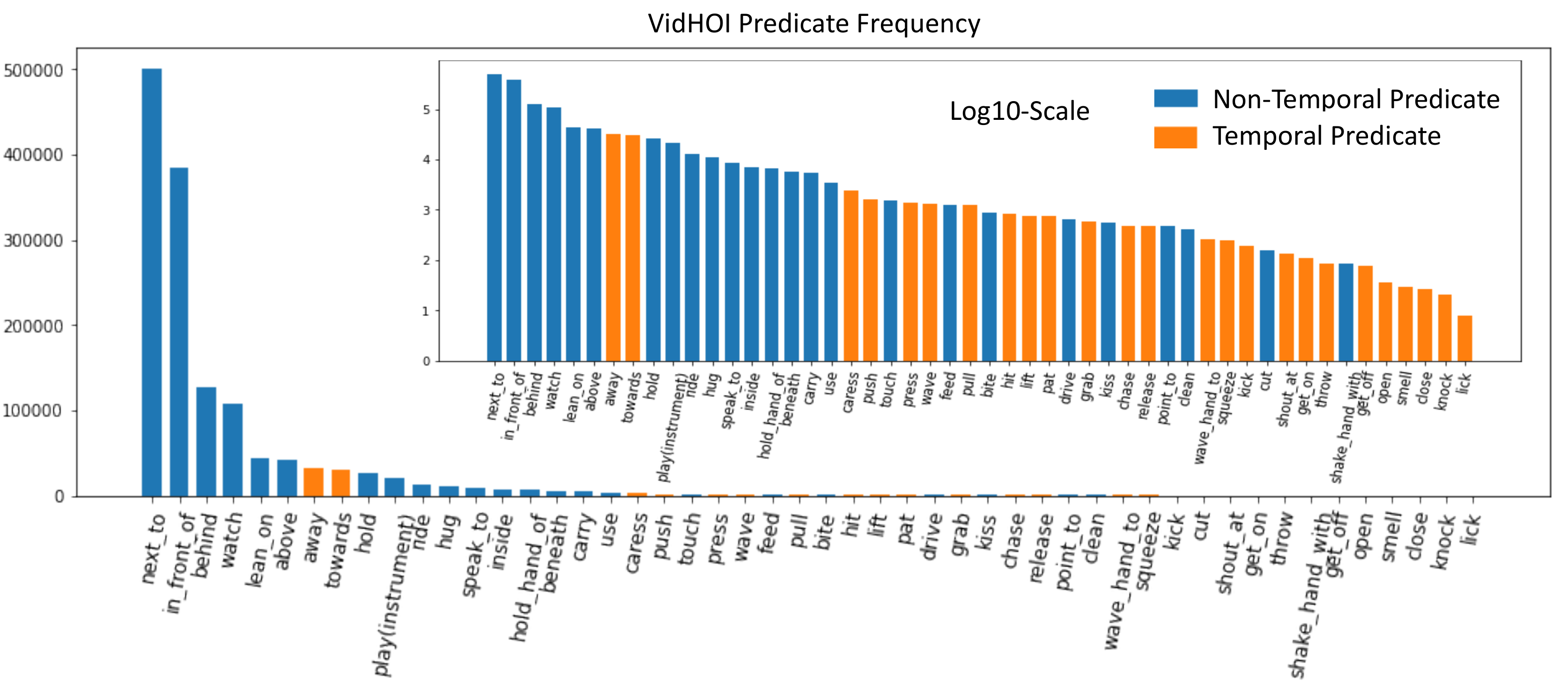}
\end{center}
\vspace{-1.3em}
   \caption{Predicate distribution of the VidHOI benchmark shows that most of the predicates are non-temporal-related.}
\label{fig:pred_dist}
\vspace{-1.5em}
\end{figure}

VidOR is originally collected for video visual relationship detection where the evaluation is trajectory-based. % and the performance is mostly bounded by underlying trajectory detectors. 
The volumetric Interaction Over Union (vIOU) between a trajectory and a ground truth needs to be over 0.5 before considering its relationship prediction; however, how to obtain accurate trajectories with correct start- and end-timestamp remains challenging \cite{sun2019video,shang2017video}.
We notice that some image-based HOI datasets (\emph{e.g.}, HICO-DET \cite{chao2018learning} and V-COCO \cite{gupta2015visual}) as well as STAD datasets (\emph{e.g.}, AVA \cite{gu2018ava}) are using a keyframe-centered evaluation strategy, which bypasses the aforementioned issue.
We thus adopt the same and follow AVA to sample keyframes at a 1 FPS frequency, where the annotations on the keyframe at timestamp $t$ are assumed to be fixed for $t \pm 0.5$s. 
In detail, we first filter out those keyframes without presenting at least one valid human-object pair, followed by transforming the labels from video clip-based to keyframe-based to align with common HOI metrics (\emph{i.e.}, frame mAP).
We follow the original VidOR split in \cite{shang2019annotating} to divide VidHOI into a training set comprising 193,911 keyframes in 6,366 videos and a validation set\footnote{The VidOR testing set is not available publicly.} with 22,808 keyframes in 756 videos. 
As shown in Figure \ref{fig:pred_dist}, there are 50 relation classes including actions (\emph{e.g.}, {\fontfamily{qcr}\selectfont push}, {\fontfamily{qcr}\selectfont pull}, {\fontfamily{qcr}\selectfont lift}, etc.) and spatial relations (\emph{e.g.}, {\fontfamily{qcr}\selectfont next to}, {\fontfamily{qcr}\selectfont behind}, etc.).
While half (25) of the predicate classes are temporal-related, they account for merely $\sim$5\% of the dataset.
% Thus, we will also analyze performance within each group.
% In VidHOI there are 78 object classes, which is two less than VidOR as we combine {\fontfamily{qcr}\selectfont adult}, {\fontfamily{qcr}\selectfont child} and {\fontfamily{qcr}\selectfont baby} into a single category {\fontfamily{qcr}\selectfont person}.

\begin{figure*}[ht!]
\centering
\includegraphics[width=0.83\textwidth]{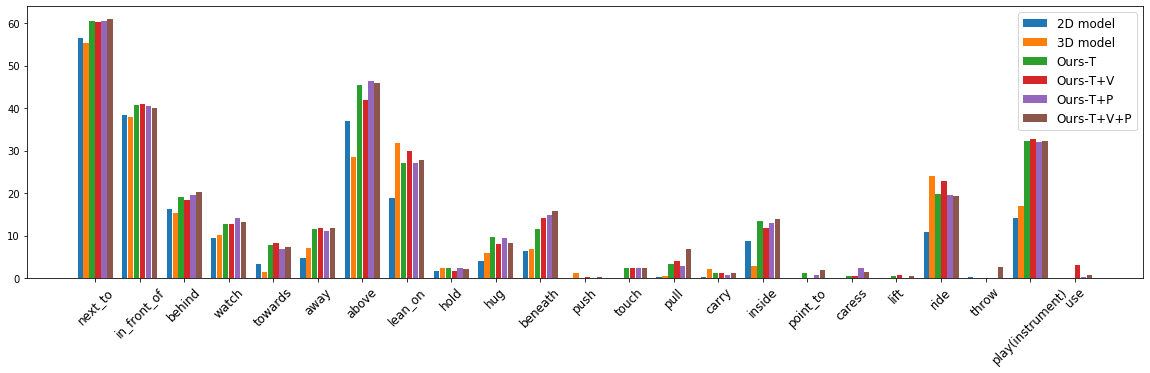}
\vspace{-1.5em}
\caption{Performance comparison in predicate-wise AP (pAP).
The performance boost after adding trajectory features is observed for most of the predicates.
Interestingly, both spatial (\emph{e.g.}, {\fontfamily{qcr}\selectfont next to}, {\fontfamily{qcr}\selectfont behind}) and temporal (\emph{e.g.}, {\fontfamily{qcr}\selectfont towards}, {\fontfamily{qcr}\selectfont away}) predicates benefit from the temporal-aware features.
% Only top performant predicate classes are shown.
Predicates sorted by the number of occurrence.
Models in {\fontfamily{qcr}\selectfont Oracle} mode.
}
\vspace{-1em}
\label{fig:predicate_map}
\end{figure*}

Following the evaluation metric in HICO-DET, we adopt mean Average Precision (mAP), where a true positive HOI needs to meet three below criteria: (a) both the predicted human and object bounding boxes have to overlap with the ground truth boxes with IOU over 0.5, (b) the predicted target category need to be matched and (c) the predicted interaction is correct. 
Over 50 predicates, we follow HICO-DET to define HOI categories as 557 triplets on which we compute mean AP.
By defining HOI categories with triplets we can bypass the polysemy problem \cite{zhong2020polysemy}, \emph{i.e.}, the same predicate word can represent very different meaning when pairing with distinct objects, \emph{e.g.}, {\fontfamily{qcr}\selectfont person-fly-kite} and {\fontfamily{qcr}\selectfont person-fly-airplane}.
We report the mean AP over three categories: (a) \textbf{Full}: all 557 categories are evaluated, (b) \textbf{Rare}: 315 categories with less than 25 instances in the dataset, and (c) \textbf{Non-rare}: 242 categories with more than or equal to 25 instances in the dataset.
We also examine the models in two evaluation modes: {\fontfamily{qcr}\selectfont Oracle} models are trained and tested with ground truth trajectories, while models in {\fontfamily{qcr}\selectfont Detection} mode are tested with predicted trajectories.

\subsection{Implementation Details}
\label{subsec:implementation}

We adopt Resnet-50 \cite{he2016deep} as our 2D backbone for the preliminary experiments, and utilize Resnet-50-based SlowFast \cite{feichtenhofer2019slowfast} as our 3D backbone for all the other experiments.
SlowFast contains the Slow and Fast pathways, which correspond to the texture details and the temporal information, respectively, by sampling video frames in different frequencies.
For a 64-frame segment centered at the keyframe, $T=32$ frames are alternately sampled to feed into the Slow pathway; only $T/\alpha$ frames are fed into the Fast pathway, where $\alpha = 8$ in our experiments. 
% We train all the models from the scratch. % since we find in preliminary experiments that finetuning on pretrained network (\emph{e.g.} SlowFast pretrained on Kinetics-400 \cite{kay2017kinetics}) does not help.
% We evaluate our method with either predicted trajectories ({\fontfamily{qcr}\selectfont Detection}) or ground truth ({\fontfamily{qcr}\selectfont Oracle}).
We use FastPose \cite{fang2017rmpe} to predict human poses and adopt the predicted trajectories generated by a cascaded model of video object detection, temporal NMS and tracking algorithm \cite{sun2019video}. 
Like object detection is to 2D HOI detection, trajectory generation is an essential module but not a main focus of this work.
If a bounding box is not available in neighboring frames (\emph{i.e.}, the trajectory is shorter than $T$ or not continuous throughout the segment), we fill it with the whole-image as a box.
We train all models from scratch for 20 epochs with the initial learning rate $1 \times 10^{-2}$, where we use step decay learning rate to reduce the learning rate by $10\times$ at the $10^{\text{th}}$ and $15^{\text{th}}$ epoch.
We optimize our models using synchronized SGD with momentum of 0.9 and weight decay of $10^{-7}$.
We train each 3D video model with eight NVIDIA Tesla V100 GPUs with batch size being 128 (\emph{i.e.}, 16 examples per GPU), except for the full model where we set batch size as 112 due to the memory restriction.
We train the 2D model with a single V100 with batch size being 128.

During training, following the strategy in SlowFast we randomly scale the shorter side of the video to a value in $[256, 320]$ pixels, followed by random horizontal flipping and random cropping into $224 \times 224$ pixels. 
During inference, we only resize the shorter side of the video segment to 224 pixels.

\subsection{Quantitative Results}
\label{subsec:quantitative}

\begin{table}
\centering
\caption{Results of the baselines and our ST-HOI on VidHOI validation set (numbers in mAP).
There are two evaluation modes: {\fontfamily{qcr}\selectfont Detection} and {\fontfamily{qcr}\selectfont Oracle}, which differ only in the use of predicted or ground truth trajectories during inference.
% The 2D model is the HOI baseline in \cite{wan2019pose} and the 3D model replaces the 2D backbone with the 3D one  \cite{feichtenhofer2019slowfast}.
\textbf{T}: Trajectory features.
\textbf{V}: Correctly-localized visual features.
\textbf{P}: Spatial-temporal masking pose features.
"\textbf{\%}" means the full mAP change compared to the 2D model.}
\vspace{-1em}
\begin{tabular}{cl|cccc}
\hline
& Model & Full & Non-rare & Rare & \% \\
\hline
\multirow{6}{*}{\rotatebox[origin=c]{90}{\textit{Oracle}}}
& 2D model \cite{wan2019pose} & 14.1 & 22.9 & 11.3 & - \\
& 3D model & 14.4 & 23.0 & 12.6 & 2.1 \\
& Ours-T & 17.3 & 26.9 & 16.8 & 22.7 \\
& Ours-T+V & 17.3 & 26.9 & 16.3 & 22.7 \\
& Ours-T+P & 17.4 & 27.1 & 16.4 & 23.4 \\
& Ours-T+V+P & \textbf{17.6} & \textbf{27.2} & \textbf{17.3} & \textbf{24.8} \\
\hline
\multirow{6}{*}{\rotatebox[origin=c]{90}{\textit{Detection}}}
& 2D model \cite{wan2019pose} & 2.6 & 4.7 & 1.7 & - \\
& 3D model & 2.6 & 4.9 & 1.9 & 0.0 \\
& Ours-T & 3.0 & 5.5 & 2.0 & 15.4 \\
& Ours-T+V & 3.1 & 5.8 & 2.0 & 19.2 \\
& Ours-T+P & \textbf{3.2} & \textbf{6.1} & 2.0 & \textbf{23.1} \\
& Ours-T+V+P & 3.1 & 5.9 & \textbf{2.1} & 19.2 \\
\hline
\vspace{-1.8em}
\label{tab:results}
\end{tabular}
\end{table}

% We mainly compare our models with simple
Since we aim to deal with a) the lack of temporal-aware features in 2D HOI methods, b) the feature inconsistency issue in common 3D HOI methods and c) the lack of a VideoHOI benchmark, we mainly compare with the 2D model \cite{wan2019pose} and its naive 3D variant on VidHOI to understand if our ST-HOI addresses these issues effectively.

The performance comparison between our full ST-HOI model (\textbf{Ours-T+V+P}) and baselines (\textbf{2D model}, \textbf{3D model}) are presented in Table \ref{tab:results}, in which we also present ablation studies on our different features (modules) inlcuding trajectory features (\textbf{T}), correctly-localized visual features (\textbf{V}) and spatial-temporal masking pose features (\textbf{P}).
Table \ref{tab:results} shows that \textbf{3D model} only has a marginal improvement compared to \textbf{2D model} (overall $\sim$2\%) under all settings in both evaluation modes.
In contrast, adding trajectory features (\textbf{Ours-T}) leads to a much larger 23\% improvement in {\fontfamily{qcr}\selectfont Oracle} mode or 15\% in {\fontfamily{qcr}\selectfont Detection} mode, showing the importance of correct spatial-temporal information.
We also find that by adding additional temporal-aware features (\emph{i.e.,} \textbf{V} and \textbf{P}) increasingly higher mAPs are attained, and our full model (\textbf{Ours-T+V+P}) reports the best mAPs in {\fontfamily{qcr}\selectfont Oracle} mode, achieving the highest $\sim$25\% relative improvement.
We notice that the performance of \textbf{Ours-T+V} is close to that of \textbf{Ours-T} under {\fontfamily{qcr}\selectfont Oracle} setting, which is possibly because the ground truth trajectories (\textbf{T}) have provided enough ``correctly-localized'' information so that the correct features do not help much.
We also note that the performance of \textbf{Ours-T+P} is slightly higher than that of \textbf{Ours-T+V+P} under {\fontfamily{qcr}\selectfont Detection} mode, which is assumably due to the same, aforementioned reason and the inferior performance resulting from the predicted trajectories.
The overall performance gap between {\fontfamily{qcr}\selectfont Detection} and {\fontfamily{qcr}\selectfont Oracle} models is significant, indicating the room for improvement in trajectory generation.
Another interesting observation is that Full mAPs are very close to Rare mAPs, especially under {\fontfamily{qcr}\selectfont Oracle} mode, showing that the long-tail effect over HOIs is strong (but common and natural).

To understand the effect of temporal features on individual predicates, we compare with predicate-wise AP (pAP) shown in Figure \ref{fig:predicate_map}.
We observe that, again, under most of circumstances naively replacing 2D backbones with 3D ones does not help video HOI detection.
Both temporal predicates (\emph{e.g.}, {\fontfamily{qcr}\selectfont towards}, {\fontfamily{qcr}\selectfont away}, {\fontfamily{qcr}\selectfont pull}) and spatial (\emph{e.g.}, {\fontfamily{qcr}\selectfont next\_to}, {\fontfamily{qcr}\selectfont behind}, {\fontfamily{qcr}\selectfont beneath}) predicates benefit from the additional temporal-aware features in ST-HOI.
These findings verify our main idea about the essential use of trajectories and trajectory-based features.
In addition, each additional features do not seem to contribute equally for different predicates.
For instance, we see that while \textbf{Ours-T+V+P} performs the best on some predicates (\emph{e.g.}, {\fontfamily{qcr}\selectfont behind} and {\fontfamily{qcr}\selectfont beneath}), our sub-models achieve the highest mAP on other predicates (\emph{e.g.}, {\fontfamily{qcr}\selectfont watch} and {\fontfamily{qcr}\selectfont ride}).
This is assumedly because predicate-wise performance is heavily subject to the number of examples, where major predicates have 10-10000 times more examples than minor ones (as shown in Figure \ref{fig:pred_dist}).

\begin{figure}[t!]
\begin{center}
% \fbox{\rule{0pt}{1in} \rule{0.9\linewidth}{0pt}}
\includegraphics[width=0.82\linewidth]{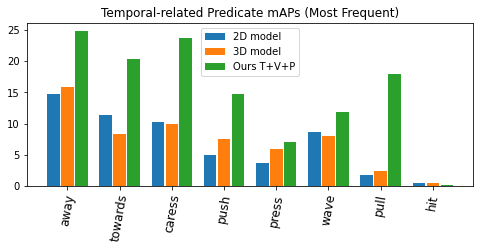}
\end{center}
\vspace{-1.3em}
\caption{Results (in predicate-wise AP) of the baselines and our full model w.r.t. top frequent temporal predicates.}
\label{fig:temp_pred_result}
\vspace{-1.5em}
\end{figure}

\begin{table}
\centering
\caption{Results of temporal-related and spatial (non-temporal) related triplet mAP. T\%/S\% means relative temporal/spatial mAP change compared to 2D model \cite{wan2019pose}.}
\vspace{-1em}
\begin{tabular}{cl|cccc}
\hline
& & Temporal & T\% & Spatial & S\% \\
\hline
\multirow{6}{*}{\rotatebox[origin=c]{90}{\textit{Oracle}}}
& 2D model \cite{wan2019pose} & 8.3 & - & 18.6 & - \\
& 3D model & 7.7 & -7.2 & 20.9 & 12.3 \\
% & Ours-T & 14.38 & 73.25 & 24.69 & 32.46 \\
& Ours-T & \textbf{14.4} & \textbf{73.5} & 24.7 & 32.8 \\
& Ours-T+V & 13.6 & 63.9 & 24.6 & 32.3 \\
& Ours-T+P & 12.9 & 55.4 & \textbf{25.0} & \textbf{34.4} \\
& Ours-T+V+P & \textbf{14.4} & \textbf{73.5} & \textbf{25.0} & \textbf{34.4}
\\
\hline
\multirow{6}{*}{\rotatebox[origin=c]{90}{\textit{Detection}}}
& 2D model \cite{wan2019pose} & 1.5 & - & 2.7 & - \\
& 3D model & 1.6 & 6.7 & 2.9 & 7.4 \\
& Ours-T & 1.8 & 20.0 & \textbf{3.3} & \textbf{23.6} \\
& Ours-T+V & 1.8 & 20.0 & \textbf{3.3} & \textbf{23.6} \\
& Ours-T+P & 1.8 & 20.0 & \textbf{3.3} & \textbf{23.6} \\
& Ours-T+V+P & \textbf{1.9} & \textbf{26.7} & \textbf{3.3} & \textbf{23.6} \\
\hline
\label{tab:temporal_spatial_results}
\end{tabular}
\vspace{-2em}
\end{table}

Since the majority of HOI examples are spatial-related ($\sim$95\%, as shown in Figure \ref{fig:pred_dist}), the results above might not be suitable for demonstrating the temporal modeling ability of our proposed model.
We thus focus on the performance on only temporal-related predicates in Figure \ref{fig:temp_pred_result}, which demonstrates that \textbf{Ours-T+V+P} greatly outperforms the baselines regarding the top frequent temporal predicates.
Table \ref{tab:temporal_spatial_results} presents the triplet mAPs of spatial- or temporal-only predicates, showing \textbf{Ours-T} significantly improves the \textbf{2D model} on temporal-only mAP by relative +73.9\%, in sharp contrast to -7.1\% of the \textbf{3D model} in {\fontfamily{qcr}\selectfont Oracle} mode.
Similar to our observation with Table \ref{tab:results}, \textbf{Ours-T} performs on par with \textbf{Ours-T+V+P} for temporal-only predicates; however, it falls short of spatial-only predicates, showing that spatial/pose information is still essential for detecting spatial predicates.
% while the trajectories have provided enough temporal information,
Overall, these results demonstrate the outstanding spatial-temporal modeling ability of our approach.

\begin{figure}[t!]
\centering
\includegraphics[width=0.85\linewidth]{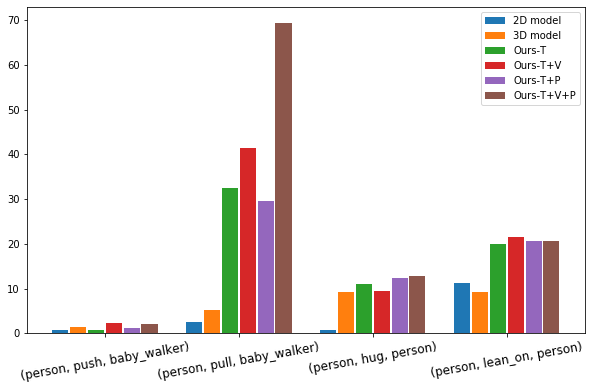}
\vspace{-1.3em}
\caption{
Performance comparison (in AP) of some temporal-related HOIs in VidHOI validation set.
% Numbers are in mAP normalized to the highest score in the six models.
Compared to 2D model, 3D model only shows limited improvement for the presented examples, while our ST-HOI variants provide huge performance boost.
% for both temporal-aware ({\fontfamily{qcr}\selectfont towards-person}, {\fontfamily{qcr}\selectfont lift-ball} and {\fontfamily{qcr}\selectfont pull-baby\_walker}) and pose-aware ({\fontfamily{qcr}\selectfont hug-person}, {\fontfamily{qcr}\selectfont lean\_on-person}) HOIs.
Models are in {\fontfamily{qcr}\selectfont Oracle} mode.
}
\vspace{-1.5em}
\label{fig:hoi_examples}
\end{figure}

\begin{figure*}[t!]
\centering
\includegraphics[width=0.78\textwidth]{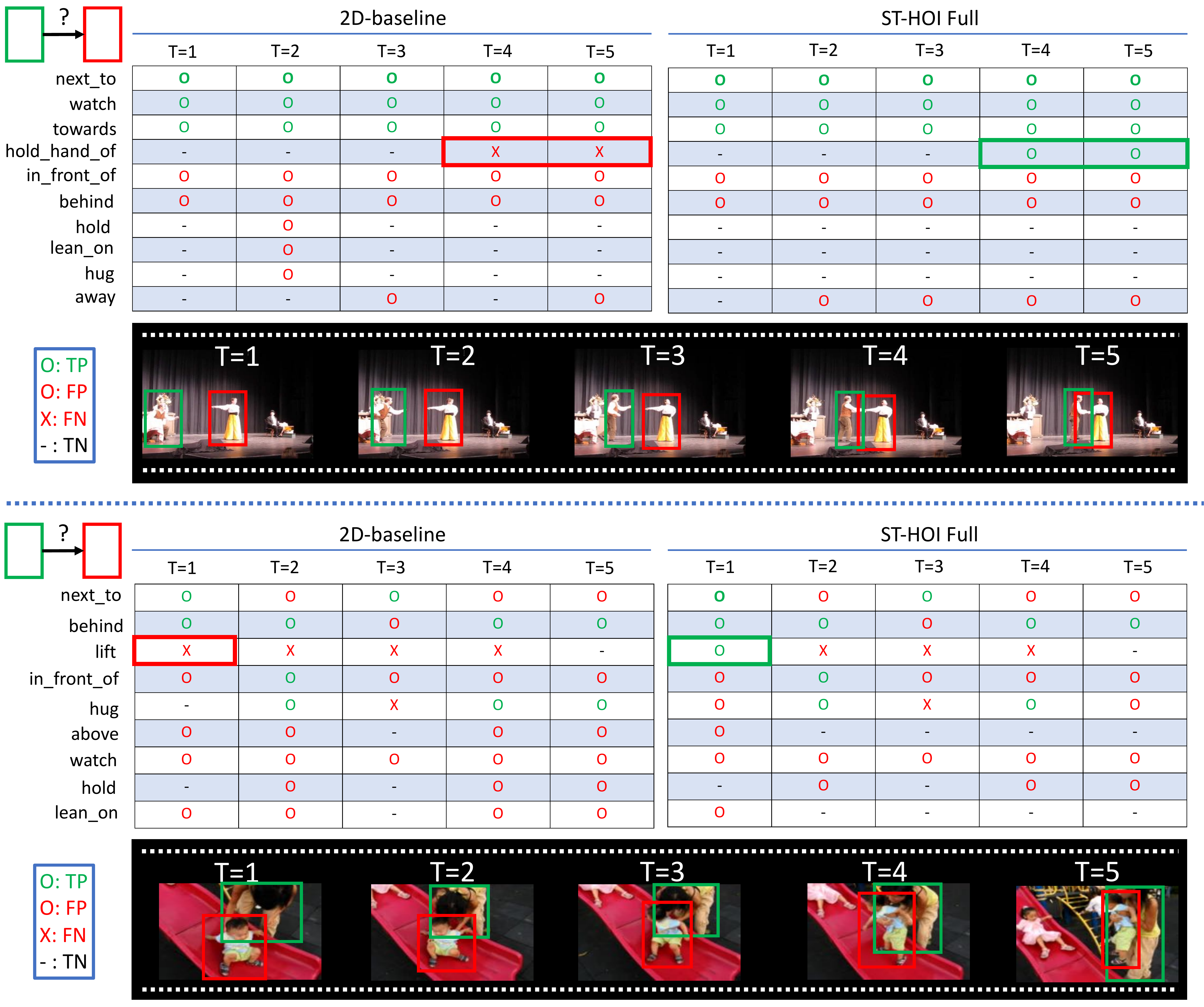}
\vspace{-1.3em}
\caption{Examples of video HOIs predicted by the 2D model \cite{wan2019pose} and our ST-HOI, both in {\fontfamily{qcr}\selectfont Oracle} mode.
Each consists of five consecutive keyframes sampled in 1 Hz, where an entry in tables denotes whether a predicate between the subject (human; a green box) and the object (also human in both cases; a red box) is detected correctly (True Positive) or not (False Positive or False Negative).
Compared to the 2D baseline, our model predicts more accurate temporal HOIs (\emph{e.g.}, {\fontfamily{qcr}\selectfont hold\_hand\_of} in $T_4$ and $T_5$ of the upper example and {\fontfamily{qcr}\selectfont lift} in $T_1$ of the lower example).
ST-HOI also produces less false positives in both examples.
}
\vspace{-1em}
\label{fig:hoi_visualization}
\end{figure*}

We also compare the performance with respect to some HOI triplets in Figure \ref{fig:hoi_examples}.
Similar to the results on predicate-wise mAP, we also observe the large gap between naive 2D/3D models and our models with the temporal features. 
ST-HOI variants are more accurate in predicting especially temporal-aware HOIs ({\fontfamily{qcr}\selectfont hug/lean\_on-person} and {\fontfamily{qcr}\selectfont push/pull-baby\_walker}).
We also see in some examples that \textbf{Ours-T+V+P} does not perform the best among all the variants, \emph{e.g.}, {\fontfamily{qcr}\selectfont lean\_on-person}), which is similar to the phenomenon we observed in Figure \ref{fig:predicate_map}.
% We suspect that the underlying reason is similar to that of we observed earlier in Figure .

% this results from the lack of training and validation examples for these HOI categories, due to the long-tail nature of the distribution.
% For example, {\fontfamily{qcr}\selectfont pull-baby\_walker} accounts for only $0.04\%$ of the total validation examples.
% As mAP is a more strict metric that also measures the ability of overcomeing long-tail distribution

\subsection{Qualitative Results}
\label{subsec:qualitative}

To understand the effectiveness of our proposed method, we visualize two video HOI examples of VidHOI predicted by the \textbf{2D model} \cite{wan2019pose} and \textbf{Ours-T+V+P} (both in {\fontfamily{qcr}\selectfont Oracle} mode) in Figure \ref{fig:hoi_visualization}.
Each (upper and lower) example is a 5-second video segment (\emph{i.e.}, five keyframes) with a HOI prediction table where each entry means either True Positive (TP), False Positive (FP), False Negative (FN) or True Negative (TN) for both models.
The upper example shows that, compared to the \textbf{2D model}, \textbf{Ours-T+V+P} makes more accurate HOI detection by successfully predicting {\fontfamily{qcr}\selectfont hold\_hand\_of} at $T_4$ and $T_5$.
Moreover, \textbf{Ours-T+V+P} is able to predict interactions that requires temporal information, such as {\fontfamily{qcr}\selectfont lift} at $T_1$ in the lower example.
However, we can see that there is still room for improvement for \textbf{Ours-T+V+P} in the same example, where {\fontfamily{qcr}\selectfont lift} is not detected in the following $T_2$ to $T_4$ frames.
Overall, our model produces less false positives throughout the dataset, which in turn contributes to its higher mAP and pAP.

%%% Conclusion %%%
\section{Conclusion}
\label{sec:conclusion}

In this paper, we addressed the inability of conventional HOI approaches to recognize temporal-aware interactions by re-focusing on neighboring video frames.
We discussed the lack of a suitable setting and dataset for studying video-based HOI detection.
We also identified a feature-inconsistency problem in a common video action detection baseline which arises from its improper order of RoI feature pooling and temporal pooling.
To deal with the first issue, we established a new video HOI benchmark dubbed VidHOI and introduced a keyframe-centered detection strategy.
We then proposed a spatial-temporal baseline ST-HOI which exploits trajectory-based temporal features including correctly-localized visual features, spatial-temporal masking pose features and trajectory features, solving the second problem.
With quantitative and qualitative experiments on VidHOI, we showed that our model provides a huge performance boost compared to both the 2D and 3D baselines and is effective in differentiating temporal-related interactions.
We expect that the proposed baseline and the dataset would serve as a solid starting point for the relatively underexplored VideoHOI task.
Based on our baseline, we also hope to motivate further VideoHOI works to design advanced models with the multi-modal data including video frames, semantic object/relation labels and audios.

\section*{Acknowledgment}
This research is supported by Singapore Ministry of Education Academic Research Fund Tier 1 under MOE's official grant number T1 251RES2029.

\bibliographystyle{ACM-Reference-Format}
\bibliography{egbib}

\end{document}